\documentclass[10pt,twocolumn,letterpaper]{article}
\pdfoutput=1

\usepackage[pagenumbers]{cvpr} 
\usepackage{graphicx}
\usepackage{amsmath}
\usepackage{amssymb}
\usepackage{booktabs}
\usepackage[accsupp]{axessibility}  

\usepackage[marginal]{footmisc}

\usepackage[pagebackref,breaklinks,colorlinks]{hyperref}

\usepackage[capitalize]{cleveref}
\crefname{section}{Sec.}{Secs.}
\Crefname{section}{Section}{Sections}
\Crefname{table}{Table}{Tables}
\crefname{table}{Tab.}{Tabs.}

\def\cvprPaperID{5231} 
\def\confName{CVPR}
\def\confYear{2022}

\begin{document}

\title{Learning Pixel-Level Distinctions for Video Highlight Detection}

\author{Fanyue Wei$^1$\footnotemark[1] \quad Biao Wang$^2$ \quad Tiezheng Ge$^2$ \quad Yuning Jiang$^2$ \quad Wen Li$^1$ \quad Lixin Duan$^1$
\vspace{3pt}\\
\normalsize$^1$School of Computer Science and Engineering \& Shenzhen Institute for Advanced Study, UESTC \quad $^2$Alibaba Group\\
\tt\small \{wfanyue, liwenbnu, lxduan\}@gmail.com, \tt\small \{eric.wb, tiezheng.gtz, mengzhu.jyn\}@alibaba-inc.com
}

\maketitle

\begin{abstract}
   The goal of video highlight detection is to select the most attractive segments from a long video to depict the most interesting parts of the video. Existing methods typically focus on modeling relationship between different video segments in order to learning a model that can assign highlight scores to these segments; however, these approaches do not explicitly consider the contextual dependency within individual segments. To this end, we propose to learn pixel-level distinctions to improve the video highlight detection. This pixel-level distinction indicates whether or not each pixel in one video belongs to an interesting section. The advantages of modeling such fine-level distinctions are two-fold. First, it allows us to exploit the temporal and spatial relations of the content in one video, since the distinction of a pixel in one frame is highly dependent on both the content before this frame and the content around this pixel in this frame. Second, learning the pixel-level distinction also gives a good explanation to the video highlight task regarding what contents in a highlight segment will be attractive to people. We design an encoder-decoder network to estimate the pixel-level distinction, in which we leverage the 3D convolutional neural networks to exploit the temporal context information, and further take advantage of the visual saliency to model the spatial distinction. State-of-the-art performance on three public benchmarks clearly validates the effectiveness of our framework for video highlight detection.
\end{abstract}

\section{Introduction}

\begin{figure}[t]
  \centering
  \includegraphics[width=\linewidth]{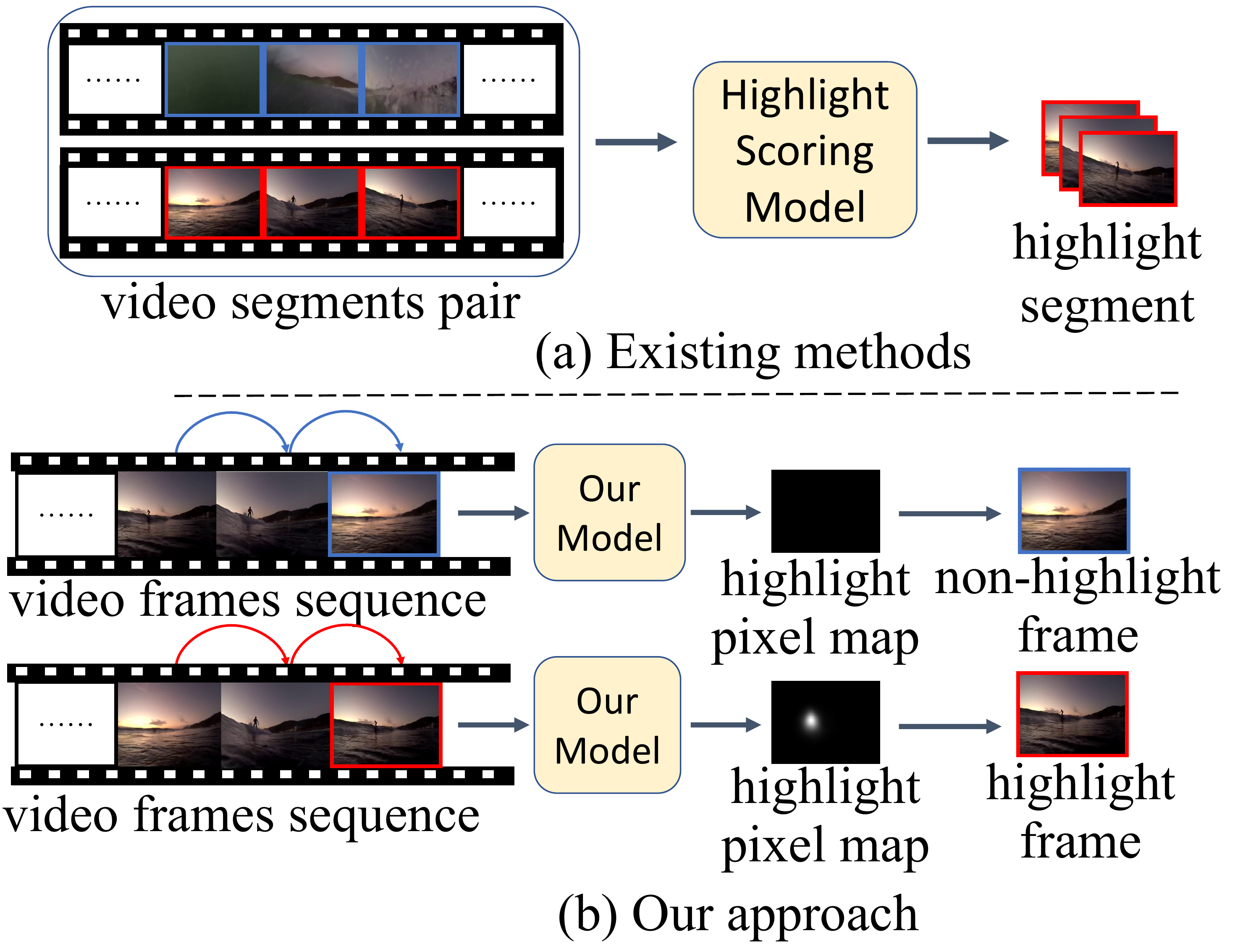}
  \caption{Video highlight detection is highly context-dependent. While previous methods are usually trained to predict the highlight score for a video segment directly, our method takes the temporal and spatial information into account and predicts the fine-level pixel-level distinction as the surrogate task.} 
\label{temporal_module}
\end{figure}

\renewcommand{\thefootnote}{\fnsymbol{footnote}}
\setlength{\footnotemargin}{2mm}
\footnotetext[1]{Work done during an internship at Alibaba Group}

Along with the explosive development of mobile devices, a tremendous number of videos are now produced and uploaded to the Internet every day. As a result, picking the most attractive video clips from a lengthy video to create a selection of shining moments is becoming increasingly important, especially for social video platforms such as YouTube and Instagram. As a result, video highlight detection, which aims to select the most attractive segments from an unedited video, has drawn increasing interest from in the research community.

Most existing works~\cite{gygli2016video2gif, yao2016highlight, jiao2018three} interpret the video highlight detection task as a segment-level ranking problem. These approaches treat each segment as an individual sample and extract the features for video segments. They then compare pairwise segments in order to learn a model that assigns highlight scores to these segments, such that the highlight segments receive higher scores than the non-highlight segments from the raw video. Recently, SL~\cite{xu2021cross} developed a set-based mechanism that is capable of identifying whether or not a video segment is highlight by transformer.

However, these existing methods do utilize both temporal and localized information, but not explicitly considering the contextual dependency within the segment, which is in fact crucial for video highlight detection. 

Intuitively, when people watch videos, a specific part is considered to be interesting, usually depends on the previous parts they have watched. For example, considering a video in which a gymnast performs a somersault, the jumping up before the somersault and jumping down after the somersault are visually quite similar; however, people tend to rate the jumping up as more appealing than the jumping down; because the former contributes to the the climax of the somersault, while the latter decays the highlight level after climax. This indicates that predicting the highlight score of one frame highly depends on the context before the current frame. 

Similarly, the spatial context is also important for video highlight detection. A dog might not be interesting if it appears together with a group of dogs, while it will definitely be the focus in a dog show scenario. In this case, the context information within one frame would be very helpful for estimating the highlight score. 

Accordingly, to exploit the temporal and spatial context for video highlight detection, in this paper, we cast the video highlight detection into a new task: pixel-level distinction estimation. More specifically, rather than assigning highlight scores to video segments (as in the existing works), we aim to predict the attractiveness of each pixel in the video. Such fine-level task offers two benefits. First, as the distinction of a pixel in one frame often depends on the temporal and spatial context, predicting pixel-level distinction allows to exploit such context information in our model, leading to more robust highlight detection results. Second, learning the pixel-level distinction also offers a good explanation for the video highlight task also to what content in a highlight segment might be more appealing to people, making the video highlight detection model more explainable. After estimating the pixel-level distinction, the highlight score of a video segment can be readily obtained by averaging the distinctions of all pixels in the segment. 

We develop an encoder-decoder network to estimate the pixel-level distinction. This network is designed to output a distinction map for each frame in the input video. To exploit the temporal context, we employ a 3D convolutional neural networks to incorporate the frames before the current frame in order to predict the distinction map. To model the spatial distinction, we take advantage of the visual saliency to generate pixel-level pseudo-distinction labels for frames in the highlight segments. We demonstrate that the strategies discussed above can be simply integrated into the encoder-decoder network. 

Experiments on three challenging benchmarks-YouTube~\cite{sun2014ranking}, TvSum~\cite{song2015tvsum} and CoSum~\cite{chu2015video}-show that our proposed approach outperforms existing methods by clear margins. We further validate the effectiveness of our proposed modules with ablation studies, and provide qualitative results to show the explainable ability of our proposed model.  

In summary, the main contributions of this paper are as follows:
\begin{itemize}
\item We propose a new pixel-level distinction estimation task for video highlight detection, which is able to explore the fine-level context in order to predict the attractiveness of specific segments. 
\item We design an encoder-decoder network for estimating the pixel-level distinction, which takes advantage of the 3D convolutional neural networks and the visual saliency map to exploit the temporal and spatial context, respectively.
\item We achieve new state-of-the-art performance on three public benchmarks. Moreover, our model also exhibits good explainable ability, and is able to directly output the most appealing regions in the highlighted video segments. 
\end{itemize}

\section{Related Work}
\subsection{Video Highlight Detection}
The goal of video highlight detection is to find the most attractive parts of videos. Prior methods have largely focused on generating highlight from sports videos~\cite{wang2004sports, xu2006live, xu2008novel, zhu2007human}. More recent approaches focus on addressing the Internet videos and first-person videos.
These recent methods can be divided into two aspects: supervised and unsupervised (or weakly supervised).

Supervised methods mainly treat video highlight detection as a segment-level ranking or scoring task~\cite{yao2016highlight, gygli2016video2gif, jiao2018three, xu2021cross, badamdorj2021joint}. These methods generally construct a pair-wise ranking constraint for two video segments, highlight and non-highlight segments. 
Video2GIF~\cite{gygli2016video2gif} proposes a method to learn from manually generated video-GIF pairs. By utilizing an adaptive Huber loss to overcome noisy data, a robust deep RankNet can generate a ranked list of video segments. 

Additionally, 
GNN~\cite{zhang2020find} introduces object semantics to video highlight task and further model the relationships between objects via graph neural network.

SL~\cite{xu2021cross} utilizes transformer structure to capture the multiple segments that contributes to the target segment. Moreover, SA~\cite{badamdorj2021joint} proposes that audio and visual information are highly related to highlight detection. They fuse audio and visual information by attention for video highlight detection.

Unsupervised or weakly supervised methods often introduces some prior information as a supervised signal rather than using the highlight annotations for training.

LIM-s~\cite{xiong2019less} leverages the video duration as the implicit supervised signal. They contend that user-generated videos have the relationship that video segments from shorter videos are more likely to be highlights than those from longer videos.
Therefore, they propose a model that learns to score the highlight segments higher than non-highlight segments. More recently, MINI-Net~\cite{hong2020mini} casts video highlight detection as multiple instance learning problem. They characterize each video as a bag of segments, aiming to score a positive bag about specific event higher than a negative bag that events are irrelevant.

Most methods generate the highlight clips by ranking the highlight and non-highlight segments based on segment-level feature representation. In a departure from the existing methods, our work captures the visual temporal distinction via sliding window and introduces visual saliency to model the highlight spatially with pixel-level loss.

\subsection{Video Summary}

The goal of video summarization, which is highly related to video highlights, is to produce the most informative clip that incorporates the complete plot of an entire given video~\cite{zhang2018retrospective, yi2019acm, otani2019rethinking, vasudevan2017query}. Video summary models often learn to score a sequence of selected frames~\cite{lee2012discovering} or clips~\cite{gygli2015video}. Additionally, some video summary methods consider not only importance but also representation~\cite{zhou2018deep}, diversity~\cite{gong2014diverse} and coherency~\cite{lu2013story}. \cite{Mahasseni_2017_CVPR} aims to select a subset of frames that optimally represent a given video, performing unsupervised video summarization with adversarial LSTM networks. The model comprises a summarizer, which aims to obtain an optimal summarization of a new video, and a discriminator, to distinguish between the original video and its reconstruction obtained from the summarizer. \cite{zhou2018deep} formalizes video summarization into a sequential decision-making process. By training an end-to-end reinforcement learning framework, the proposed model  predicts frame-level probability to be chosen to form summary. Moreover, some methods ~\cite{zhao2017hierarchical, zhao2018hsa} take a hierarchical recurrent neural network to exploit the long temporal dependency among frames for video summarization. \cite{zhao2021reconstructive} captures the temporal dependencies with LSTM and GCN hierarchically.

\subsection{Visual Saliency}
Visual saliency aims to model the gaze fixation. Previous methods have utilized optical flow to make use of temporal information~\cite{wang2015consistent, bak2017spatio}. Moreover, some methods aggregate temporal information with LSTM~\cite{Bazzani:ICLR2017}. ACLNet~\cite{wang2018revisiting} enhances the ability of LSTM to capture the dynamic saliency through the use of a frame-wise attention mask. 
Recently, TASED-Net~\cite{min2019tased} aggregates temporal information and takes an encoder-decoder structure spatially to predict frame-wise saliency maps in a sliding-window fashion for a given video. STAViS~\cite{tsiami2020stavis} combines spatiotemporal auditory and visual information to address video saliency.

\section{Approach}

In this paper, we propose to leverage the temporal and spatial relations within a segment to improve video highlight detection. Our motivation derives from the fact that video highlights is highly context-dependent; \ie, whether or not the content in a video segment should be highlighted depends on the content that comes before it in temporal dimension and the content surrounding it in the spatial dimension.

Paradigms in previous methods~\cite{sun2014ranking, gygli2016video2gif, zhang2020find, xu2021cross, badamdorj2021joint} have proposed different strategies for learning the scoring function $f(s_{i})$, which largely involve assigning higher scores to highlight segments and lower scores to non-highlight segments. These methods usually obtain the whole feature representation from each video segment, and learn score function differently. However, these methods tend to ignore the spatial-temporal relation of the content among frames within each segment, which is in fact crucial to video highlight detection.

To capture this context-dependent property, rather than scoring each video segment on its global feature representation, we instead propose to predict the highlight score for each pixel per frame, an approach referred to as \emph{\textbf{P}ixel-\textbf{L}evel \textbf{D}istinction \textbf{V}ideo \textbf{H}ighlight \textbf{D}etection (\textbf{PLD-VHD})} in the paper.

\begin{figure*}[t]
  \centering
  \includegraphics[width=\linewidth]{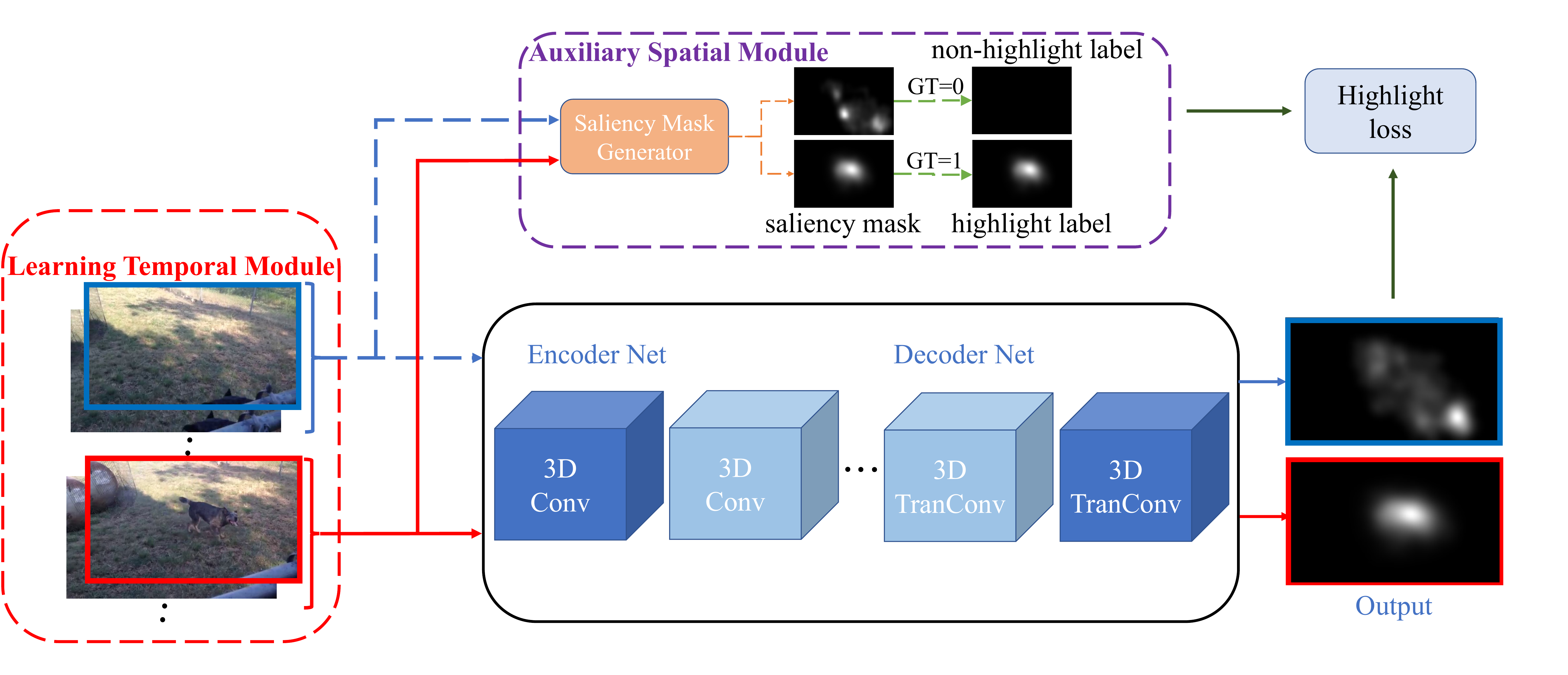}
  \caption{Our network follows an encoder-decoder structure. The encoder net is a 3D ConvNet for extracting features of input frames, while the decoder net aims to obtain a map that is the same size as the input frames for pixel-level distinction. The learning temporal module incorporates the previous frames before the target frame, while the auxiliary spatial module generates the pseudo labels. The frames in blue and red boxes represent targets with two kinds of labels: highlight and non-highlight, respectively.}
  \label{network_structure}
\end{figure*}

\subsection{Modeling Temporal Dependency}

Formally, for one video $V$, denote $S = \{ s_{1},\ldots, s_{n}\}$ as the set of video segments after division, where each $s_{i}$ is a segment for $i = 1, \ldots, n$. Each $s_{i}$ is accompanied by a label $y_{i}$, where $y_{i}=1$ indicates that $s_{i}$ is a highlight segment while $y_{i}=0$ denotes opposite.

We begin from a basic model for pixel-level distinction estimation. As the ground-truth pixel-level distinction $d_t(i,j)$ is unknown, we need to construct a pseudo-distinction label for each pixel using the segment-level highlight label $y_s$. The basic concept of our approach is simple. For frames from non-highlight segments $s_{n}$, we set the distinction labels of all pixels as zeros; for those from highlight segments $s_{h}$, their distinction labels are set as ones. The pseudo-distinction label can be defined as follows:

\begin{equation}
d_t(i,j) = \left\{
\begin{array}{rcl}
1, & & I_{t} \in s_{h} \\
0, & & I_{t} \in s_{n}
\end{array} \right.
\label{all_01_label}
\end{equation}

For simplicity, we use $D_t$ to represent the distinction map for a frame $I_t$, where $d_t(i,j)$ is the pseudo-distinction label defined in Eq.~(\ref{all_01_label}). For ease of presentation, we also use $f(I_t)$ to denote the pixel-level distinction estimation function for the entire frame $I_t$. We then take a simple mean squared error (MSE) as the loss, and the problem for learning pixel-level distinctions can be formulated as:
\begin{equation}
\min \mathcal{L}(f(I_{t}), D_{t}) = \frac{1}{W \cdot H} \sum_{i=1}^{W} \sum_{j=1}^{H}  \left( p_t(i,j) - d_t(i,j)\right)^{2}
\label{mse_loss}
\end{equation}
where $f(\cdot)$ is the distinction estimation function, $p_{t}(i, j)$ denotes the pixel-level distinctions obtained by $f(\cdot)$, $W$ and $H$ denote the width and height of the frame. 

The distinction estimation function can be implemented with an encoder-decoder network. The input frame $I_{t}$ are first fed into an encoder to obtain the latent feature representation, after which the feature map is upsampled by the decoder network for pixel-level distinction prediction. 

However, as discussed above, creating video highlights is highly context-dependent. When people watch videos, the current frame becomes interesting because people have watched the previous frames. This means that distinctions of the current frame should depend on the frames that came before it. 

Accordingly, to take the temporal dependency into account, rather than directly using each frame as input, we use a video clip to predict the pixel-level distinction. This clip contains both the current frame and a number of frames before it. Given the $t$-th frame $I_t$, let us denote the corresponding video clip as $C_t = \{I_{t-L+1}, I_{t-L+2}, \ldots, I_{t}\}$ where $L$ is the total length of the video clip. The model for predicting the pixel-level distinctions can thus be updated as follows:
\begin{equation}
    \min \mathcal{L}(f(C_{t}), D_{t})
\label{mse_clip_loss}
\end{equation}
where $\mathcal{L}$ is the MSE loss defined as in Eq.~(\ref{mse_loss}).

In our implementation, we apply an $L$-length sliding window to the video to generate the video clips. For the first $T-1$ frames in each video, we reverse the order of these frames and pad them to the beginning of the video to ensure the sliding window works. Each video clip is then fed into a 3D convolutional neural network (\eg, C3D~\cite{tran2015learning} or TASED~\cite{min2019tased}) to automatically exploit the temporal relation among frames within each video clip, as illustrated in Figure~\ref{temporal_module}.

\begin{table*}[t]
\centering
\caption{Video highlight detection results of different methods on the YouTube Highlights dataset. }
\label{compared_methods}
\begin{tabular}{c|ccccccccc||c}    
\hline
           & LSVM  & RRAE & Video2gif  & LIM-s  & MINI-Net & AFM-F-M  & GNN & SL & SA & PLD-VHD\\
\hline
dog        & 0.60  & 0.49 & 0.308 & 0.579  & 0.582   & 0.72 & 0.67 & 0.708 & 0.649 & \textbf{0.749} \\
gymnastics & 0.41  & 0.35 & 0.335 & 0.417  & 0.617   & 0.56 & 0.66 & 0.532 & \textbf{0.715} & 0.702 \\
parkour    & 0.61  & 0.50 & 0.540 & 0.670  & 0.702   & 0.75 & \textbf{0.83} & 0.772 & 0.766 & 0.779  \\
skating    & 0.62  & 0.25 & 0.554 & 0.578  & 0.722   & 0.68 & 0.70 & \textbf{0.725} & 0.606 & 0.575 \\
skiing     & 0.36  & 0.22 & 0.328 & 0.486  & 0.587   & 0.64 & 0.69 & 0.661 & \textbf{0.712} & 0.707 \\
surfing    & 0.61  & 0.49 & 0.541 & 0.651  & 0.651   & 0.78 & 0.69 & 0.762 & 0.782 & \textbf{0.790} \\
\hline
Average    & 0.536 & 0.412& 0.464 & 0.564  & 0.644   & 0.68 & 0.69 & 0.693 & 0.705 & \textbf{0.730}  \\
\hline
\end{tabular}
\label{results_youtube}
\end{table*}

\begin{table*}[t]
\centering
\caption{Experimental results (top-5 mAP score) of compared methods on the TVsum dataset.}
\small
\setlength{\tabcolsep}{1.2mm}{
\begin{tabular}{c|c|c|c|c|c|c|c|c|c|c|c|c|c|c||c}
\hline
                       & KVS   & DPP   & sLstm & SM  & Quasi & MBF   & CVS   & SG    & LIM-s & VESD  & DSN   & MINI-Net & SL & SA & PLD-VHD       \\
\hline
BK                     & 0.342 & 0.395 & 0.406 & 0.407 & 0.295 & 0.313 & 0.326 & 0.417 & 0.663 & 0.441 & 0.368 & 0.717            & 0.726 & 0.681 & \textbf{0.845}     \\
BT                     & 0.419 & 0.464 & 0.471 & 0.473 & 0.327 & 0.365 & 0.402 & 0.483 & 0.691 & 0.492 & 0.435 & 0.769            & 0.789 & \textbf{0.950} & 0.809    \\
DS                     & 0.394 & 0.449 & 0.455 & 0.453 & 0.309 & 0.357 & 0.378 & 0.466 & 0.626 & 0.488 & 0.416 & 0.591           & 0.640 & 0.608 & \textbf{0.703}     \\
FM                     & 0.397 & 0.442 & 0.452 & 0.451 & 0.318 & 0.365 & 0.365 & 0.464 & 0.432 & 0.487 & 0.412 & 0.559           & 0.589 & 0.669 & \textbf{0.725}     \\
GA                     & 0.402 & 0.457 & 0.463 & 0.469 & 0.342 & 0.325 & 0.379 & 0.475 & 0.612 & 0.496 & 0.428 & 0.754            & 0.749 & \textbf{0.844} & 0.764     \\
MS                     & 0.417 & 0.462 & 0.477 & 0.478 & 0.375 & 0.412 & 0.398 & 0.489 & 0.54  & 0.503 & 0.436 & 0.813            & 0.862 & 0.865 & \textbf{0.872}     \\
PK                     & 0.382 & 0.437 & 0.448 & 0.445 & 0.324 & 0.318 & 0.354 & 0.456 & 0.604 & 0.478 & 0.411 & 0.780   & \textbf{0.790} & 0.703 & 0.719\\
PR                     & 0.403 & 0.446 & 0.461 & 0.458 & 0.301 & 0.334 & 0.381 & 0.473 & 0.475 & 0.485 & 0.417 & 0.545            & 0.632 & 0.675 & \textbf{0.740}     \\
VT                     & 0.353 & 0.399 & 0.411 & 0.415 & 0.336 & 0.295 & 0.328 & 0.423 & 0.559 & 0.447 & 0.373 & 0.803    & \textbf{0.865} & 0.834 & 0.744     \\
VU                     & 0.441 & 0.453 & 0.462 & 0.467 & 0.369 & 0.357 & 0.413 & 0.472 & 0.429 & 0.493 & 0.441 & 0.653            & 0.687 & 0.647 & \textbf{0.791}     \\
\hline
Average                & 0.398 & 0.447 & 0.451 & 0.461 & 0.329 & 0.345 & 0.372 & 0.462 & 0.563 & 0.481 & 0.424 & 0.698            & 0.733 & 0.748 & \textbf{0.771}     \\
\hline
\end{tabular}
}
\label{results_tvsum}
\end{table*}

\begin{table*}

\small
\caption{Results (top-5 mAP score) on the CoSum Dataset. Our method outperforms all comparison methods by a large margin.}
\centering
\setlength{\tabcolsep}{1.8mm}{
\begin{tabular}{c|c|c|c|c|c|c|c|c|c|c|c|c||c}

\hline
    & KVS & DPP & sLstm & SM & SMRS & Quasi & MBF & CVS & SG & VESD & DSN & MINI-Net & PLD-VHD \\ 
\hline
BJ & 0.662 & 0.672 & 0.683 & 0.692 & 0.504 & 0.561 & 0.631 & 0.658 & 0.698 & 0.685 & 0.715 & 0.776 & \textbf{0.900} \\
BP & 0.674 & 0.682 & 0.701 & 0.722 & 0.492 & 0.625 & 0.592 & 0.675 & 0.713 & 0.714 & 0.746 & 0.963 & \textbf{0.970}   \\
ET & 0.731 & 0.744 & 0.749 & 0.789 & 0.556 & 0.575 & 0.618 & 0.722 & 0.759 & 0.783 & 0.813 & 0.786 & \textbf{0.817}   \\
ERC & 0.685 & 0.694 & 0.717 & 0.728 & 0.525 & 0.563 & 0.575 & 0.693 & 0.729 & 0.721 & 0.756 & 0.953 & \textbf{1.000} \\
KP & 0.701 & 0.705 & 0.714 & 0.745 & 0.521 & 0.557 & 0.594 & 0.707 & 0.729 & 0.742 & 0.772 & 0.959 & \textbf{1.000} \\ 
MLB & 0.668 & 0.677 & 0.714 & 0.693 & 0.543 & 0.563 & 0.624 & 0.679 & 0.721 & 0.687 & 0.727 & 0.869 & \textbf{1.000} \\ 
NFL & 0.671 & 0.681 & 0.681 & 0.727 & 0.558 & 0.587 & 0.603 & 0.674 & 0.693 & 0.724 & 0.737 & 0.897 & \textbf{0.970} \\ 
NDC & 0.698 & 0.704 & 0.722 & 0.759 & 0.496 & 0.617 & 0.694 & 0.702 & 0.738 & 0.751 & 0.782 & 0.890 & \textbf{0.958} \\ 
SL & 0.713 & 0.722 & 0.721 & 0.766 & 0.525 & 0.551 & 0.624 & 0.715 & 0.743 & 0.763 & 0.794 & 0.787 & \textbf{0.844} \\ 
SF & 0.642 & 0.648 & 0.653 & 0.653 & 0.533 & 0.562 & 0.603 & 0.647 & 0.681 & 0.674 & 0.709 & 0.727 & \textbf{1.000} \\
\hline
Average & 0.684 & 0.692 & 0.705 & 0.735 & 0.525 & 0.576 & 0.602 & 0.687 & 0.720 & 0.721 & 0.755 & 0.861 & \textbf{0.946} \\
\hline
\end{tabular}
}
\label{results_cosum}
\end{table*}

\begin{figure}[t]
  \centering
  \includegraphics[width=\linewidth]{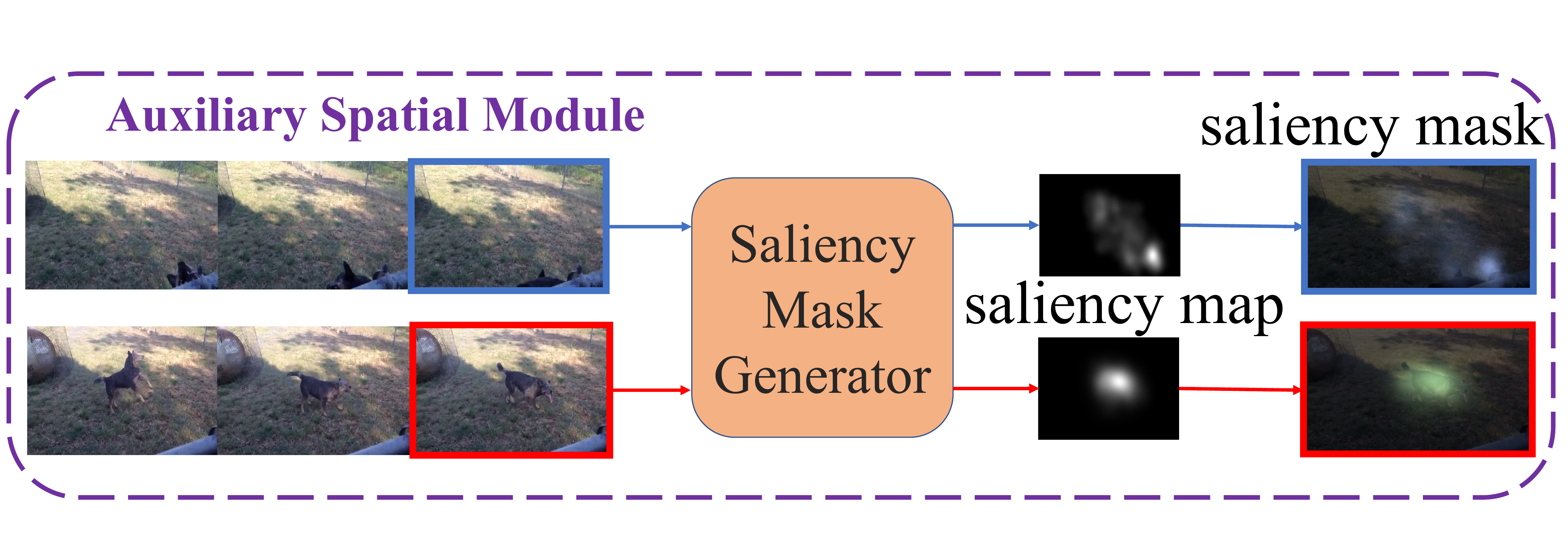}
  \caption{The spatial distinction of our approach. With the mask generated by our Saliency Mask Generator, the pseudo-label can eliminate the noise (such as the background) that makes no contribution to the highlight.}
  \label{spatial_module}
\end{figure}

\subsection{Spatial Highlight with Visual Saliency}

In addition to the contextual temporal dependency within video clips, the attractiveness of an object is also often dependent on the surrounding context. For example, a single dog might not be particularly appealing when it appears in a group of dogs in one image, but it is the shining star in a dog show scenario. 

We therefore further consider the spatial relationship for learning pixel-level distinctions. 
In particular, for non-highlight segments, their pseudo-labels are still all-zeros, since none of the pixels in these segments are of interest. For highlight segments, we take advantage of the visual saliency to exploit the spatial context in each frame.

On one hand, as Figure~\ref{spatial_module} shows, visual saliency, which can be seen as robust general visual signals, aims to model the gaze fixation people display when they are watching videos, which is in line with the goal of video highlight detection. Using saliency helps us to identify the fine-level regions that attract people. On the other hand, while we annotate the pixel-level distinctions for all pixels in the highlight segments in Eq.~(\ref{all_01_label}), not all regions in the highlight segments are truly attractive, which yields a considerable amount of noise when optimizing the learning problem in Eq.~(\ref{mse_clip_loss}). Using saliency information to eliminate the background noise facilitates the learning of a more robust pixel-level distinction estimation model. 

More specifically, we use the saliency mask as the pseudo-labels to annotate the pixel-level distinctions for pixels in the highlight segments. Given any frame $I_t$ in the highlight segment, we denote its saliency mask as $M_{t}$. The pixel-level distinctions can thus be defined as follows:
\begin{equation}
    \hat{d}_{t}(i,j)=\left\{
\begin{array}{rcl}
0, &  & { M_{t}(i, j) \leq \beta }\\
1, &  & { M_{t}(i, j) > \beta}
\end{array} \right.
\label{mask_label}
\end{equation}
where $\beta$ is a hyper-parameter threshold we simply set as 0.0005 in most cases.

Note that, by using the above definition of pixel-level distinctions $\hat{d}_{t}(i,j)$ to replace the original pixel-level distinctions ${d}_{t}(i,j)$ in Eq.~(\ref{all_01_label}), the spatial distinction can be seamlessly integrated with the temporal dependency learning framework. We are also able to jointly exploit the spatial and temporal dependencies in order to estimate the pixel-level distinctions. Denoting $\hat{D}_t$ as the new distinction map for frame $I_t$, the learning objective can be updated as follows:
\begin{equation}
    \min \mathcal{L} (f(C_t), \hat{D}_t) 
\label{final_loss}
\end{equation}
where $\mathcal{L}$ is the MSE loss defined as in Eq.(\ref{mse_loss}).

After learning the pixel-level distinction estimation model, given a segment from any video, the highlight score can be calculated by averaging all pixel-level distinctions in the segments, as follows:
\begin{equation}
    f(s_{d}) = \sum_{t=1}^{N} \sum_{i=1}^{H} \sum_{j=1}^{W} \frac{1}{N \cdot H \cdot W} (f(C_t)^{(i,j)})
\end{equation}
where $s_{d}$ denotes the $d$-th segment in a video, while $f(C_t)^{(i,j)}$ is the $(i,j)$-th element of the estimated distinction map for $I_t$; moreover, $N, H$ and $W$ respectively denote the number of frames in $s_d$, the height and the width of the frames. The highlight score of a video segment can be estimated by using the mean of all pixel-level distinctions in this segment, while the highlighted video can be obtained by ensembling the video segments with highest scores similarly as in existing video highlight detection works.

\subsection{Network Architecture}

As shown in Figure~\ref{network_structure}, our model is constituted by an encoder network (used to extract the features of the input video clip) and a decoder network (used to generate a distinction map corresponding to the target frame). The \emph{Temporal Module} is designed to obtain the auxiliary past information for the current frame to be predicted. It utilizes 3D convolution neural network input consisting of past consequent frames and the current target frame. Moreover, the \emph{Auxiliary Spatial Module} is a visual saliency model that works as an encoder network to generate a saliency mask. For this purpose, we here adopt TASED-Net~\cite{min2019tased} pretrained on DHF1K~\cite{wang2018revisiting}. The final output of our whole framework is a highlight map as same size as the input target frame. 

\begin{figure*}[t]
  \centering
  \includegraphics[width=\linewidth]{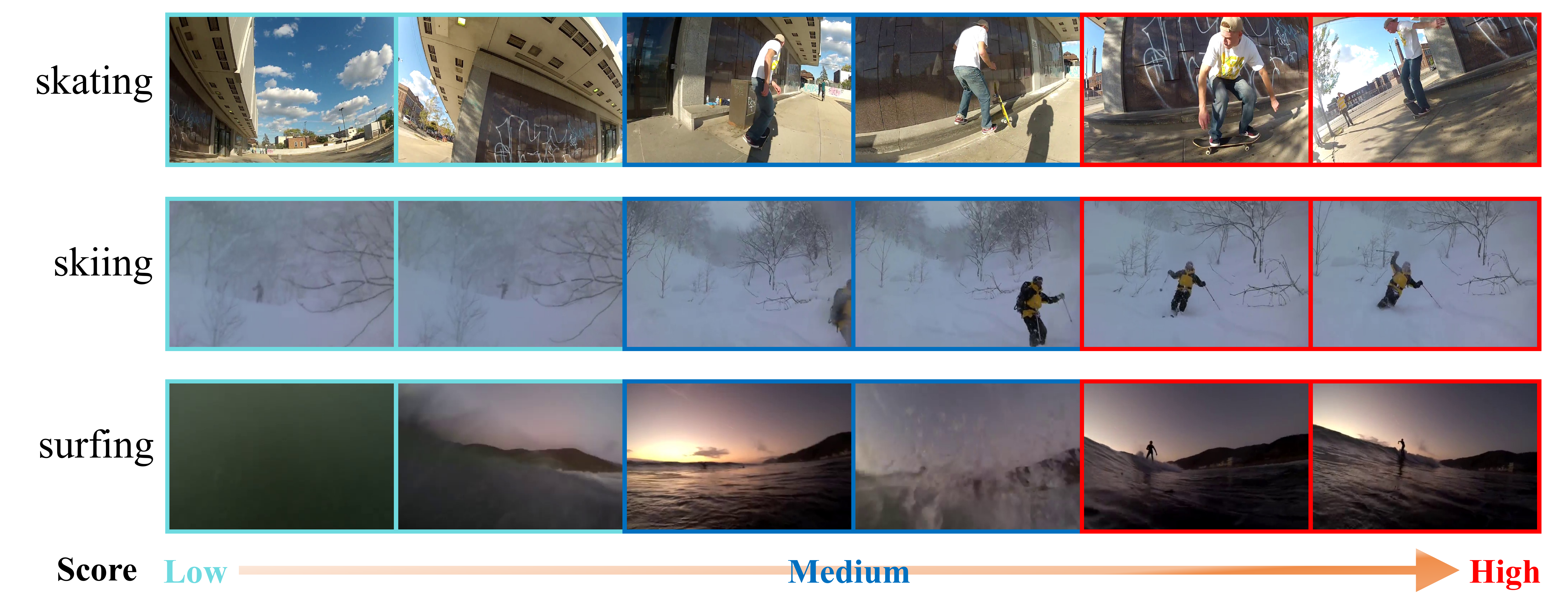}
  \caption{Showcase examples from different domain. Red means a higher highlight score, light green indicates a lower highlight score, and blue represents a medium highlight score.}
\label{showcase}
\end{figure*}

\begin{table*}[t]
\centering
\caption{Results of ablation studies on YouTube Highlights dataset.}
\setlength{\tabcolsep}{1.2mm}{
\begin{tabular}{c|ccc|ccc}    
\hline
              & C3D w/o temporal   & C3D w/o spatial & C3D full  & TASED w/o temporal  &  TASED w/o spatial  & TASED full \\
\hline
dog           & 0.594   & 0.700    & 0.718         & 0.668             & 0.734        & 0.749 \\
gymnastics    & 0.707   & 0.716    & 0.730         & 0.691             & 0.701        & 0.702 \\
parkour       & 0.578   & 0.677    & 0.746         & 0.658             & 0.756        & 0.779 \\
skating       & 0.360   & 0.405    & 0.490         & 0.411             & 0.521        & 0.575 \\
skiing        & 0.667   & 0.670    & 0.696         & 0.654             & 0.705        & 0.707 \\
surfing       & 0.725   & 0.756    & 0.758         & 0.736             & 0.779        & 0.790 \\
\hline
Average       & 0.651   & 0.664    & 0.712         & 0.667             & 0.702        & 0.730  \\
  
\hline
\end{tabular}
}
\label{ablation_youtube}
\end{table*}

\section{Experiments}

In this section, we validate our model on several challenging public benchmarks-YouTube~\cite{sun2014ranking}, TvSum~\cite{song2015tvsum} and CoSum~\cite{chu2015video}-and compare the results with those of several state-of-the-art video highlight detection methods. More experimental details are reported in the Supplementary Materials.

\subsection{Experimental Setup}

\subsubsection{Dataset and Evaluation Metric}
\begin{itemize}
    \item \emph{YouTube Highlight}~\cite{sun2014ranking} is a popular video highlight detection dataset that collects videos from six different domains. Each domain contains 50 to 90 videos with varying duration. Each video is divided into several segments that contain approximately 100 frames, each of which are annotated with three different kinds of labels: 1-selected by users as highlight; 0-borderline cases, and -1–non-highlight. We treat the borderline cases as non-highlight. 
    \item \emph{TvSum}~\cite{song2015tvsum} contains 50 videos of 10 classes. Following~\cite{xiong2019less, hong2020mini}  we select top 50\% of shots in terms of the scores provided by annotators for each video as the human-created ground truth.
    \item \emph{CoSum}~\cite{chu2015video} consists of 51 videos of 10 events. In this work, following~\cite{hong2020mini}, we compare each generated highlight with three human-created ground truth.
\end{itemize}

Like most existing methods~\cite{xiong2019less, hong2020mini}, we follow Video2gif~\cite{gygli2016video2gif}, using mean average precision(\textbf{mAP}) as the evaluation metric.

\subsubsection{Comparison Methods}
We compare our methods (\emph{PLD-VHD}) with the following state-of-the-art video highlight detection baselines on three datasets.
\begin{itemize}
    \item \emph{Weakly supervised methods.} The comparison methods in this category are RRAE~\cite{yang2015unsupervised}, MBF~\cite{chu2015video}, SMRS~\cite{smrselhamifar2012see}, Quasi~\cite{quasikim2014joint}, CVS~\cite{cvspanda2017collaborative}, SG~\cite{Mahasseni_2017_CVPR}, VESD~\cite{vesdcai2018weakly}, DSN~\cite{dsnpanda2017weakly}, LIM-s~\cite{xiong2019less} and MINI-Net~\cite{hong2020mini}. 
    \item \emph{Supervised methods.} There are also several supervised methods selected for comparison, \ie, LSVM~\cite{sun2014ranking}, Video2gif~\cite{gygli2016video2gif}, KVS~\cite{kvspotapov2014category}, DPP~\cite{gong2014diverse}, sLstm~\cite{slstmzhang2016video}, SM~\cite{gygli2015video}, AFM-F-M~\cite{jiao2018three}, GNN~\cite{zhang2020find}, SL~\cite{xu2021cross} and SA~\cite{badamdorj2021joint}. 
\end{itemize}
Although some of these methods are used for video summarization, following~\cite{xiong2019less, hong2020mini}, their performance is evaluated using the same metrics as those used in this study.

\subsection{Video Highlight Detection Results}

The public datasets contain videos under different situations, such as camera view changes. In terms of the overall experimental results, our proposed method with pseudo-distinction labels achieves the best performance.
Table~\ref{results_youtube} presents the results of different methods for video highlight detection on the \emph{YouTube Highlight dataset}~\cite{sun2014ranking}. We report the results of our proposed approach using TASED-Net~\cite{min2019tased} as the backbone networks, denoted by ``\textbf{\emph{PLD-VHD}}". For the baseline methods, their results are copied from their original papers or borrowed from~\cite{xiong2019less, hong2020mini}.

Moreover, \emph{PLD-VHD} improves the SOTA methods (\ie SA~\cite{badamdorj2021joint} and MINI-Net~\cite{hong2020mini}) on \emph{TvSum} and \emph{CoSum} by 3.1\% and 9.9\%, respectively, as shown in Tables~\ref{results_tvsum} and~\ref{results_cosum}, respectively.  
This clearly demonstrates the effectiveness of our approach by learning pixel-level distinctions. For limitations, our method mainly fails in first-person videos shot by hand-held cameras, especially for skating in YouTube Highlight, which contain a lot of cluttered background due to uncontrollable camera motions

We also present some visual examples for video highlight detection in different domains. 

As it is shown in Figure~\ref{showcase} and the supplementary material, by learning pixel-level distinctions, our framework can perform video highlight detection effectively. 

\subsection{Ablation Studies}

\begin{figure*}[ht]
  \centering
  \includegraphics[width=\linewidth]{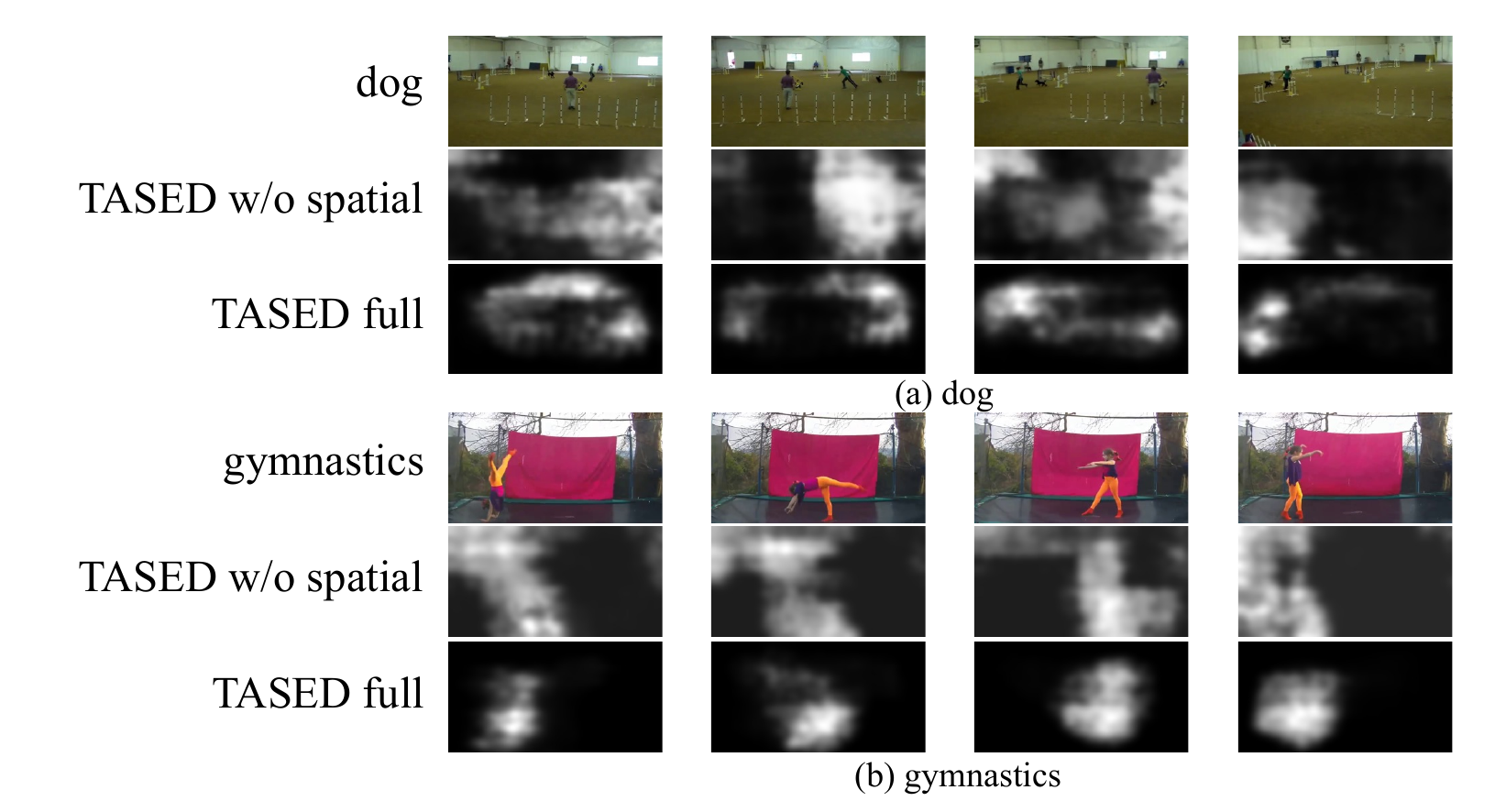}
  \caption{The frames on the first line of each subfigure are sampled from dog(gymnastics) in YouTube Highlight dataset. In the third line, the white regions inferred by our full model present the trajectory of the dog show and the action of the actress in the gymnastics clip, while the second line inferenced without the spatial module may contain some background noise and cannot provide a clear highlight cues.}
\label{highlight_map}
\end{figure*}

\begin{table}[ht]
\centering
\small
\caption{Results of ablation studies on TvSum and CoSum.}
\setlength{\tabcolsep}{1mm}{
\begin{tabular}{c|ccc}    
\hline
              & TASED w/o temporal  & TASED w/o spatial  & TASED full \\
\hline
TvSum       & 0.729    & 0.741        & 0.771 \\
CoSum       & 0.888    & 0.915        & 0.946 \\
\hline
\end{tabular}
}
\label{ablation_tvsum_cosum}
\end{table}

We conducted an additional experiment by changing the backbone network to C3D~\cite{tran2015learning} pretrained on Sports-1M~\cite{karpathy2014large}, which is the same setting using in video2gif, and is simpler than that~\cite{hara2018can} used in MINI-Net. As is shown in Table~\ref{ablation_youtube}, although the performance of Ours(C3D) is slightly worse than Ours(TASED) due to the use of a weaker backbone networks, it still outperforms all other existing methods. This again confirms that our proposed approach is effective even when different backbone networks are used.

We further validate the effectiveness of different components in our proposed approach. Specifically, two kinds of cues are used in our approach: the temporal cues and spatial cues. To validate their effects, we conduct ablation experiments by respectively removing those two types of cues as outlined below, with a total of four variants: 
\begin{itemize}
    \item \textbf{C3D w/o spatial} takes C3D~\cite{tran2015learning} as the backbone network, but does not use the saliency mask to generate the pseudo-distinction labels. In other words, we use the distinction label defined in Eq.~(\ref{all_01_label})  in this case. 
    \item \textbf{C3D w/o temporal} removes the effect of the temporal context by duplicating the target frame $I_t$ to fill the video clip $C_t$. It can be determined using only $I_t$ for distinction estimation as described in  Eq.~(\ref{mask_label}). 
    \item \textbf{TASED w/o spatial} follows the same setting as \emph{C3D w/o spatial}, but uses the TASED-Net as the backbone. 
    \item \textbf{TASED w/o temporal} follows the same setting as \emph{C3D w/o temporal}, but uses the TASED-Net as the backbone. 
\end{itemize}

The results of different variants and the full models are shown in Table~\ref{ablation_youtube}. We can observe from these results that both temporal and spatial cues are important. In particular, when spatial cues are removed, the performance of our model using C3D (TASED) drops from 0.712 to 0.664 (0.730 to 0.702). Benefiting from the spatial highlight cues, our model can be more robust to noise and exhibit improved learning of pixel-level distinction for video highlight detection, as shown in Figure~\ref{highlight_map}. Similarly, when the temporal context is eliminated, our model using C3D (TASED) drops from 0.712 to 0.651 (0.730 to 0.667). Similar ablation studies of our temporal and spatial module on TvSum and CoSum are presented in Table~\ref{ablation_tvsum_cosum}.

More detailed results of each domain on \emph{TvSum} and \emph{CoSum} are included in the supplementary materials. This confirms our analysis suggesting that the video highlight task is highly dependent on the context preceding the current frame, both temporally and spatially. 

\section{Conclusion}
In this work, we make pixel-level distinctions for video highlight detection by exploiting the temporal and spatial relations within video segments. For temporal relations, we utilize a 3D convolutional neural network to capture the distinctions by incorporating frames prior to the current frame while also making use of visual saliency to model the distinctions for spatial relations. We further adopt an encoder-decoder structure to predict pixel-level distinctions for highlight detection. In addition to achieving state-of-the-art performance, our proposed approach also has the advantage of explanability.

\section*{Acknowledgements}
This work is supported by the Major Project for New Generation of AI under Grant No. 2018AAA0100400, the National Natural Science Foundation of China (Grant No. 62176047), Beijing Natural Science Foundation (Z190023), and Alibaba Group through Alibaba Innovation Research Program.

\newpage

{\small
\bibliographystyle{ieee_fullname}
\bibliography{egbib}

\begin{thebibliography}{10}
\providecommand{\url}[1]{#1}
\csname url@samestyle\endcsname
\providecommand{\newblock}{\relax}
\providecommand{\bibinfo}[2]{#2}
\providecommand{\BIBentrySTDinterwordspacing}{\spaceskip=0pt\relax}
\providecommand{\BIBentryALTinterwordstretchfactor}{4}
\providecommand{\BIBentryALTinterwordspacing}{\spaceskip=\fontdimen2\font plus
\BIBentryALTinterwordstretchfactor\fontdimen3\font minus
  \fontdimen4\font\relax}
\providecommand{\BIBforeignlanguage}[2]{{%
\expandafter\ifx\csname l@#1\endcsname\relax
\typeout{** WARNING: IEEEtran.bst: No hyphenation pattern has been}%
\typeout{** loaded for the language `#1'. Using the pattern for}%
\typeout{** the default language instead.}%
\else
\language=\csname l@#1\endcsname
\fi
#2}}
\providecommand{\BIBdecl}{\relax}
\BIBdecl

\bibitem{hong2020mini}
F.-T. Hong, X.~Huang, W.-H. Li, and W.-S. Zheng, ``Mini-net: Multiple instance
  ranking network for video highlight detection,'' in \emph{ECCV}, 2020, pp.
  345--360.

\bibitem{xiong2019less}
B.~Xiong, Y.~Kalantidis, D.~Ghadiyaram, and K.~Grauman, ``Less is more:
  Learning highlight detection from video duration,'' in \emph{CVPR}, 2019, pp.
  1258--1267.

\bibitem{gygli2016video2gif}
M.~Gygli, Y.~Song, and L.~Cao, ``Video2gif: Automatic generation of animated
  gifs from video,'' in \emph{CVPR}, 2016, pp. 1001--1009.

\bibitem{yao2016highlight}
T.~Yao, T.~Mei, and Y.~Rui, ``Highlight detection with pairwise deep ranking
  for first-person video summarization,'' in \emph{CVPR}, 2016, pp. 982--990.

\bibitem{zhang2020find}
Y.~Zhang, J.~Gao, X.~Yang, C.~Liu, Y.~Li, and C.~Xu, ``Find objects and focus
  on highlights: Mining object semantics for video highlight detection via
  graph neural networks,'' in \emph{AAAI}, 2020, pp. 12\,902--12\,909.

\bibitem{garcia2018phd}
A.~Garcia~del Molino and M.~Gygli, ``Phd-gifs: personalized highlight detection
  for automatic gif creation,'' in \emph{ACM MM}, 2018, pp. 600--608.

\bibitem{rochan2020adaptive}
M.~Rochan, M.~K.~K. Reddy, L.~Ye, and Y.~Wang, ``Adaptive video highlight
  detection by learning from user history,'' in \emph{ECCV}, 2020, pp.
  261--278.

\bibitem{sun2014ranking}
M.~Sun, A.~Farhadi, and S.~Seitz, ``Ranking domain-specific highlights by
  analyzing edited videos,'' in \emph{ECCV}, 2014, pp. 787--802.

\bibitem{yang2015unsupervised}
H.~Yang, B.~Wang, S.~Lin, D.~Wipf, M.~Guo, and B.~Guo, ``Unsupervised
  extraction of video highlights via robust recurrent auto-encoders,'' in
  \emph{ICCV}, 2015, pp. 4633--4641.

\bibitem{jiao2018three}
Y.~Jiao, Z.~Li, S.~Huang, X.~Yang, B.~Liu, and T.~Zhang, ``Three-dimensional
  attention-based deep ranking model for video highlight detection,''
  \emph{IEEE Transactions on Multimedia}, vol.~20, no.~10, pp. 2693--2705,
  2018.

\bibitem{wang2004sports}
J.~Wang, C.~Xu, E.~Chng, and Q.~Tian, ``Sports highlight detection from keyword
  sequences using hmm,'' in \emph{ICME}, vol.~1, 2004, pp. 599--602.

\bibitem{xu2006live}
C.~Xu, J.~Wang, K.~Wan, Y.~Li, and L.~Duan, ``Live sports event detection based
  on broadcast video and web-casting text,'' in \emph{ACM MM}, 2006, pp.
  221--230.

\bibitem{xu2008novel}
C.~Xu, J.~Wang, H.~Lu, and Y.~Zhang, ``A novel framework for semantic
  annotation and personalized retrieval of sports video,'' \emph{IEEE
  transactions on multimedia}, vol.~10, no.~3, pp. 421--436, 2008.

\bibitem{qi2020emotion}
F.~Qi, X.~Yang, and C.~Xu, ``Emotion knowledge driven video highlight
  detection,'' \emph{IEEE Transactions on Multimedia}, 2020.

\bibitem{song2015tvsum}
Y.~Song, J.~Vallmitjana, A.~Stent, and A.~Jaimes, ``Tvsum: Summarizing web
  videos using titles,'' in \emph{CVPR}, 2015, pp. 5179--5187.

\bibitem{chu2015video}
W.-S. Chu, Y.~Song, and A.~Jaimes, ``Video co-summarization: Video
  summarization by visual co-occurrence,'' in \emph{CVPR}, 2015, pp.
  3584--3592.

\bibitem{zhu2007human}
G.~Zhu, Q.~Huang, C.~Xu, L.~Xing, W.~Gao, and H.~Yao, ``Human behavior analysis
  for highlight ranking in broadcast racket sports video,'' \emph{IEEE
  Transactions on Multimedia}, vol.~9, no.~6, pp. 1167--1182, 2007.

\bibitem{guo2021taohighlight}
Z.~Guo, Z.~Zhao, W.~Jin, W.~Dazhou, L.~Ruitao, and J.~Yu, ``Taohighlight:
  Commodity-aware multi-modal video highlight detection in e-commerce,''
  \emph{IEEE Transactions on Multimedia}, 2021.

\bibitem{xu2021cross}
M.~Xu, H.~Wang, B.~Ni, R.~Zhu, Z.~Sun, and C.~Wang, ``Cross-category video
  highlight detection via set-based learning,'' in \emph{ICCV}, 2021, pp.
  7970--7979.

\bibitem{badamdorj2021joint}
T.~Badamdorj, M.~Rochan, Y.~Wang, and L.~Cheng, ``Joint visual and audio
  learning for video highlight detection,'' in \emph{ICCV}, 2021, pp.
  8127--8137.

\bibitem{gong2014diverse}
B.~Gong, W.-L. Chao, K.~Grauman, and F.~Sha, ``Diverse sequential subset
  selection for supervised video summarization,'' \emph{NIPS}, vol.~27, pp.
  2069--2077, 2014.

\bibitem{zhou2018deep}
K.~Zhou, Y.~Qiao, and T.~Xiang, ``Deep reinforcement learning for unsupervised
  video summarization with diversity-representativeness reward,'' in
  \emph{AAAI}, 2018.

\bibitem{lee2012discovering}
Y.~J. Lee, J.~Ghosh, and K.~Grauman, ``Discovering important people and objects
  for egocentric video summarization,'' in \emph{CVPR}, 2012, pp. 1346--1353.

\bibitem{gygli2015video}
M.~Gygli, H.~Grabner, and L.~Van~Gool, ``Video summarization by learning
  submodular mixtures of objectives,'' in \emph{CVPR}, 2015, pp. 3090--3098.

\bibitem{lu2013story}
Z.~Lu and K.~Grauman, ``Story-driven summarization for egocentric video,'' in
  \emph{CVPR}, 2013, pp. 2714--2721.

\bibitem{Mahasseni_2017_CVPR}
B.~Mahasseni, M.~Lam, and S.~Todorovic, ``Unsupervised video summarization with
  adversarial lstm networks,'' in \emph{CVPR}, July 2017.

\bibitem{zhao2017hierarchical}
B.~Zhao, X.~Li, and X.~Lu, ``Hierarchical recurrent neural network for video
  summarization,'' in \emph{ACM MM}, 2017, pp. 863--871.

\bibitem{zhao2018hsa}
------, ``Hsa-rnn: Hierarchical structure-adaptive rnn for video
  summarization,'' in \emph{CVPR}, 2018, pp. 7405--7414.

\bibitem{zhao2021reconstructive}
B.~Zhao, H.~Li, X.~Lu, and X.~Li, ``Reconstructive sequence-graph network for
  video summarization,'' \emph{IEEE Transactions on Pattern Analysis and
  Machine Intelligence}, 2021.

\bibitem{zhang2018retrospective}
K.~Zhang, K.~Grauman, and F.~Sha, ``Retrospective encoders for video
  summarization,'' in \emph{ECCV}, 2018, pp. 383--399.

\bibitem{yi2019acm}
G.~Yi, D.~Yang, A.~Bentaleb, W.~Li, Y.~Li, K.~Zheng, J.~Liu, W.~T. Ooi, and
  Y.~Cui, ``The acm multimedia 2019 live video streaming grand challenge,'' in
  \emph{ACM MM}, 2019, pp. 2622--2626.

\bibitem{otani2019rethinking}
M.~Otani, Y.~Nakashima, E.~Rahtu, and J.~Heikkila, ``Rethinking the evaluation
  of video summaries,'' in \emph{CVPR}, 2019, pp. 7596--7604.

\bibitem{vasudevan2017query}
A.~B. Vasudevan, M.~Gygli, A.~Volokitin, and L.~Van~Gool, ``Query-adaptive
  video summarization via quality-aware relevance estimation,'' in \emph{ACM
  MM}, 2017, pp. 582--590.

\bibitem{tran2015learning}
D.~Tran, L.~Bourdev, R.~Fergus, L.~Torresani, and M.~Paluri, ``Learning
  spatiotemporal features with 3d convolutional networks,'' in \emph{ICCV},
  2015, pp. 4489--4497.

\bibitem{carreira2017quo}
J.~Carreira and A.~Zisserman, ``Quo vadis, action recognition? a new model and
  the kinetics dataset,'' in \emph{CVPR}, 2017, pp. 6299--6308.

\bibitem{hara2018can}
K.~Hara, H.~Kataoka, and Y.~Satoh, ``Can spatiotemporal 3d cnns retrace the
  history of 2d cnns and imagenet?'' in \emph{CVPR}, 2018, pp. 6546--6555.

\bibitem{xie2018rethinking}
S.~Xie, C.~Sun, J.~Huang, Z.~Tu, and K.~Murphy, ``Rethinking spatiotemporal
  feature learning: Speed-accuracy trade-offs in video classification,'' in
  \emph{ECCV}, 2018, pp. 305--321.

\bibitem{min2019tased}
K.~Min and J.~J. Corso, ``Tased-net: Temporally-aggregating spatial
  encoder-decoder network for video saliency detection,'' in \emph{ICCV}, 2019,
  pp. 2394--2403.

\bibitem{li2020two}
J.~Li, H.~Zhang, W.~Wan, and J.~Sun, ``Two-class 3d-cnn classifiers combination
  for video copy detection,'' \emph{Multimedia Tools and Applications},
  vol.~79, no.~7, pp. 4749--4761, 2020.

\bibitem{chen2018fast}
Z.~Chen, D.~Huang, Y.~Wang, and L.~Chen, ``Fast and light manifold cnn based 3d
  facial expression recognition across pose variations,'' in \emph{ACM MM},
  2018, pp. 229--238.

\bibitem{niemirepo2020binocular}
T.~T. Niemirepo, M.~Viitanen, and J.~Vanne, ``Binocular multi-cnn system for
  real-time 3d pose estimation,'' in \emph{ACM MM}, 2020, pp. 4553--4555.

\bibitem{hao2020person}
Y.~Hao, Z.-N. Liu, H.~Zhang, B.~Zhu, J.~Chen, Y.-G. Jiang, and C.-W. Ngo,
  ``Person-level action recognition in complex events via tsd-tsm networks,''
  in \emph{ACM MM}, 2020, pp. 4699--4702.

\bibitem{karpathy2014large}
A.~Karpathy, G.~Toderici, S.~Shetty, T.~Leung, R.~Sukthankar, and L.~Fei-Fei,
  ``Large-scale video classification with convolutional neural networks,'' in
  \emph{CVPR}, 2014, pp. 1725--1732.

\bibitem{wu2020learning}
L.~Wu, Y.~Yang, L.~Chen, D.~Lian, R.~Hong, and M.~Wang, ``Learning to transfer
  graph embeddings for inductive graph based recommendation,'' in \emph{SIGIR},
  2020, pp. 1211--1220.

\bibitem{wang2015consistent}
W.~Wang, J.~Shen, and L.~Shao, ``Consistent video saliency using local gradient
  flow optimization and global refinement,'' \emph{IEEE Transactions on Image
  Processing}, vol.~24, no.~11, pp. 4185--4196, 2015.

\bibitem{bak2017spatio}
C.~Bak, A.~Kocak, E.~Erdem, and A.~Erdem, ``Spatio-temporal saliency networks
  for dynamic saliency prediction,'' \emph{IEEE Transactions on Multimedia},
  vol.~20, no.~7, pp. 1688--1698, 2017.

\bibitem{Bazzani:ICLR2017}
L.~Bazzani, H.~Larochelle, and L.~Torresani, ``Recurrent mixture density
  network for spatiotemporal visual attention,'' in \emph{ICLR}, 2017.

\bibitem{wang2018revisiting}
W.~Wang, J.~Shen, F.~Guo, M.-M. Cheng, and A.~Borji, ``Revisiting video
  saliency: A large-scale benchmark and a new model,'' in \emph{CVPR}, 2018,
  pp. 4894--4903.

\bibitem{tsiami2020stavis}
A.~Tsiami, P.~Koutras, and P.~Maragos, ``Stavis: Spatio-temporal audiovisual
  saliency network,'' in \emph{CVPR}, 2020, pp. 4766--4776.

\bibitem{smrselhamifar2012see}
E.~Elhamifar, G.~Sapiro, and R.~Vidal, ``See all by looking at a few: Sparse
  modeling for finding representative objects,'' in \emph{CVPR}, 2012, pp.
  1600--1607.

\bibitem{quasikim2014joint}
G.~Kim, L.~Sigal, and E.~P. Xing, ``Joint summarization of large-scale
  collections of web images and videos for storyline reconstruction,'' in
  \emph{CVPR}, 2014, pp. 4225--4232.

\bibitem{cvspanda2017collaborative}
R.~Panda and A.~K. Roy-Chowdhury, ``Collaborative summarization of
  topic-related videos,'' in \emph{CVPR}, 2017, pp. 7083--7092.

\bibitem{vesdcai2018weakly}
S.~Cai, W.~Zuo, L.~S. Davis, and L.~Zhang, ``Weakly-supervised video
  summarization using variational encoder-decoder and web prior,'' in
  \emph{ECCV}, 2018, pp. 184--200.

\bibitem{dsnpanda2017weakly}
R.~Panda, A.~Das, Z.~Wu, J.~Ernst, and A.~K. Roy-Chowdhury, ``Weakly supervised
  summarization of web videos,'' in \emph{ICCV}, 2017, pp. 3657--3666.

\bibitem{kvspotapov2014category}
D.~Potapov, M.~Douze, Z.~Harchaoui, and C.~Schmid, ``Category-specific video
  summarization,'' in \emph{ECCV}, 2014, pp. 540--555.

\bibitem{slstmzhang2016video}
K.~Zhang, W.-L. Chao, F.~Sha, and K.~Grauman, ``Video summarization with long
  short-term memory,'' in \emph{ECCV}, 2016, pp. 766--782.

\end{thebibliography}
}

\end{document}